\pdfoutput=1
\documentclass[11pt]{article}

\usepackage[]{acl}

\usepackage{times}
\usepackage{latexsym}

\usepackage[T1]{fontenc}
\usepackage[utf8]{inputenc}

\usepackage{microtype}

\usepackage{amsmath}
\usepackage{amssymb}

\title{EigenNoise: A Contrastive Prior to Warm-Start Representations}

\author{Hunter Heidenreich \\
  Harvard University \\
  Department of Computer Science\\
  \texttt{hheidenreich@g.harvard.edu} \\\And
  Jake Ryland Williams \\
  Drexel University \\
Department of Information Science\\
  \texttt{jw3477@drexel.edu} \\}

\begin{document}

\maketitle

\begin{abstract}
    In this work, 
    we present a na\"ive initialization scheme
    for word vectors
    based on a dense, independent 
    co-occurrence model 
    and provide preliminary results
    that suggests it is competitive,
    and warrants further investigation.
    Specifically,
    we demonstrate through information-theoretic minimum description length (MDL) probing
    that our model, 
    EigenNoise,
    %eigenGloVe,
    can approach the performance 
    of empirically trained GloVe
    despite the lack of \textit{any} pre-training data
    (in the case of EigenNoise).
    %eigenGloVe.
    We present these preliminary results
    with interest to set the stage 
    for further investigations into 
    how this competitive initialization
    works without pre-training data, 
    as well as to invite the exploration 
    of more intelligent initialization schemes
    informed by the theory 
    of harmonic linguistic structure. 
    Our application of this theory likewise contributes 
    a novel (and effective) interpretation
    of recent discoveries which have elucidated the 
    underlying distributional information that
    linguistic representations capture
    from data and contrast distributions.
\end{abstract}

\section{Introduction}

Within the last decade, 
representation learning in NLP 
has experienced many major shifts,
from context-independent word vectors \cite{mikolov2013efficient,mikolov2013distributed,pennington2014glove},
to context-dependent word representations \cite{howard2018universal,peters2018deep},
to pre-trained language models \cite{devlin2019bert,radford2018improving,radford2019language}. 
These trends have been accompanied by
large architectural developments
from the dominance of RNNs \cite{hochreiter1997lstm}, 
to the appearance of attention \cite{bahdanau2015neural}
and the proliferation of the Transformer architecture \cite{vaswani2017attention}.

Despite gains on empirical benchmarks,
recent works suggest surprising findings:
word order may not matter as much in pre-training as previously thought \cite{sinha2021masked},
random sentence encodings are surprisingly powerful \cite{wieting2018no},
one can replace self-attention operations in BERT \cite{devlin2019bert}
with unparameterized Fourier transformations and still retain 92\% of the original accuracy on GLUE \cite{lee2021fnet},
and many modifications to the Transformer architecture 
do not significantly impact model performance \cite{narang2021transformer}.
While there's no denying increases in empirical performance,
these confounding results indicate % seem to indicate that 
a lack of understanding of these models
and the processing needed to perform NLP tasks.

In this work,
we take a step back and 
consider the (slightly older, yet still popular)
paradigm of 
context independent word vector algorithms
like GloVe and word2vec. 
Specifically, we reflect on
the relationships between 
prediction-based, neural methods 
and co-occurrence matrix factorization,
proposing a naive model of co-occurrence
which assumes all words co-occur at least once.
Such a naive assumption yields
a co-occurrence matrix that can be directly
computed and used as a representation
for words based on their rank-frequency,
and we provide preliminary results 
that indicate that such an approach
is surprisingly competitive 
to an empirically trained model.

\section{Background}

\subsection{Word Vectors as Matrix Factorization}

There is a deep connection between word representation algorithms
and factorization of co-occurrence matrices. 
This is transparent in GloVe \cite{pennington2014glove} by definition, 
as the log-co-occurrence counts are factored in an online fashion by minimizing Eq.~\ref{eq:GloVe-loss}, with word vectors $u,v$, bias parameters $a,b$, and $f$, a weighting function:
\begin{equation}
% \mathcal{L} =
\sum_{i,j}f(X_{ij})\left(\vec{u}_i\vec{v}_j^T + a_i + b_j - \log{X_{ij}}\right)^2
\label{eq:GloVe-loss}
\end{equation}
Similarly, word2vec's skipgram with negative sampling (SGNS) \cite{mikolov2013efficient}
has been shown to implicitly factor a co-occurrence distribution's shifted pointwise mutual information (PMI) matrix \cite{levy2014neural}, namely: $\vec{u}_i\vec{v}_j^T\approx \log\frac{X_{ij}M}{x_iy_jk}$,
%$\text{SPPMI}_k(w,c) = \text{max}{(\text{PMI}(w, c) - \log{k}, 0)}$ 
where $k$ is the number of negative samples, $M$ is the total number of co-occurrences, and $x_i$ and $y_i$ are the marginal number of co-occurrences for the $i^\text{th}$ row and $j^\text{th}$ column.
Critically, word2vec's negative samples 
lead its vectors to factor a matrix 
that provides relative information
% This is an interesting matrix to factor
about \textit{how independently} words co-occur \cite{levy2015improving,salle2019so}. 

Some suggest % recent work suggest that
that contrast helps improve quality,
especially for rare words and syntax % -aware representations 
\cite{salle2019so,shazeer2016swivel}.
Word2vec supposedly differs from GloVe's strict absorption of positive co-occurrences. However, we now know that GloVe's bias vectors seemingly each model $X$'s marginal distributions independently~\cite{kenyon-dean-etal-2020-deconstructing}. Specifically, GloVe's bias terms appear to optimize as $a_i \approx \log{x_i}$ and $b_j \approx \log{y_j}$.
So while some researchers have noted that
% differentiates our approach to that of Swivel \cite{shazeer2016swivel},
%which notes 
GloVe is \emph{under-defined} by only training on positive observations
% and imposes a penalty for spurious association of non-co-occurring words
~\cite{shazeer2016swivel}, 
we now know that GloVe's bias terms essentially learn the missing contrastive information during optimization~\cite{kenyon-dean-etal-2020-deconstructing}.
\textit{This means GloVe is roughly equivalent to SGNS-word2vec}. Specifically, while SGNS-word2vec is granted its contrastive information via marginal sampling, GloVe na\"ively utilizes bias parameters, which optimize towards the same marginal contrast distributions, independently. This connection has taken time to emerge from the literature, and from it now we ask a research question that is core to this work: how effective is a representation learned from contrastive information, alone?

% factoring the log co-occurrence matrix, however,
% However, our model assumes that every word in a vocabulary will co-occur at least once,
% allowing downstream learning to prune un-informative connections.
% This differentiates our approach to that of Swivel \cite{shazeer2016swivel},
% which notes that GloVe is \emph{under-defined} by only training on positive observations
% and imposes a penalty for spurious association of non-co-occurring words.
% Our approach circumvents this issue as every type in our vocab
% has positive co-occurrence. % in our idealized model.

\subsection{Evaluating Representations via Probes}

% There is an ever-growing body of literature 
% that attempts to understand what information is captured 
Significant work has gone into understanding the information captured
in language representations \cite{clark2019bert,conneau2018you,hewittliang2019designing,tenney2019bert,vigbelinkov2019analyzing,voita2019analyzing}.
Early work centered on
intrinsic and extrinsic properties \cite{schnabel2015evaluation},
the differences between dense and count-based vectors \cite{baroni2014don,levy2015improving},
and the information contained in a single sentence vector \cite{conneau2018you}.
% With %growing model complexity and 
Transitioning towards large pre-trained language models
has shifted focus towards characterizing %attempting to characterize
% and understand 
what these models are learning 
to understand the linguistic phenomena captured 
within learned representations \cite{hewittliang2019designing}
and self-attention maps \cite{clark2019bert,tenney2019bert,vigbelinkov2019analyzing,voita2019analyzing}.

% Recently, many of these studies focus on probing a representation
% by measuring the accuracy that a classifier may achieve on a task,
% when given data encoded with that representation
One method of understanding relies on probing a representation
by measuring classifier accuracy enabled with a representation
\cite{hewittliang2019designing,zhangbowman2018language}. 
However, many approaches fail to sufficiently differentiate 
the properties of learned representations \cite{voita2020information}.
This is especially apparent with the high performance of random baselines \cite{wieting2018no,zhangbowman2018language} 
and the ability of probes to accurately encode random labels \cite{hewittliang2019designing}.
% Some have proposed better methods for evaluation 
% by restricting probes to ``simple'' architectures \cite{hewittliang2019designing},
% presenting scores in the context of what a probe can achieve 
% on randomly assigned labels \cite{hewittliang2019designing},
% or restricting the amount of data a probe is allowed to train on 
% \cite{zhangbowman2018language}.

\section{Effective Word Vectors, Sans Data}

\subsection{Contrast and Co-occurrence}
Since LMs can seemingly learn from shuffled data and retain a surprising amount of predictive power~\cite{sinha2021masked}, 
it appears that 
% we now know that 
a great deal of information exists in contrastive information on its own. In the context of co-occurrences $X$, learning from shuffled data is equivalent to learning from independent (co-occurrence) statistics, e.g., via the cross product of $X$'s marginals. While we seek to determine the extent to which independent statistics are behind the predictive power of deep learning algorithms for benchmark NLP applications, we note that PMI \textit{must} be constant-zero on independently-occurring joint distributions (by PMI's definition). Hence, we cannot simply study independent models of data through the lens of standard GloVe or SGNS-word2vec, leading us to exclude bias terms from GloVe. When paired with a model, $\hat{X}$, of independent co-occurrences, this is roughly equivalent to learning an SGNS-word2vec model via strictly contrastive learning information. 

Removing GloVe's bias terms also simplifies its analysis. This is further aided by relieving GloVe of its weighting function:
\begin{equation}
% \mathcal{L} =
\sum_{i,j}\left(\vec{u}_i\vec{v}_j^T - \log{X_{ij}}\right)^2
\label{eq:GloVe-loss_adj}
\end{equation}
and has the effect of de-biasing optimization by row, i.e., un-balancing the learning rates that GloVe had modulated for lower-frequency words in its formulation.
This simple form allows us to straightforwardly approach the word embeddings' common objective's underlying matrix-factorization problem, whose analytic solution requires $\min_{ij}\{\hat{X}_{ij}\}>0$. In other words, provided all word pairs are modeled to co-occur at least once, the loss can easily be solved in closed forms by well-known matrix factorizations, e.g., by an eigen-decomposition. While the positivity of $\hat{X}$ can be ensured without assuming independence, another immediate benefit of studying contrastive (independent) co-occurrence models is the guarantee that they provide for $\hat{X}$'s positivity. Specifically, since $x_i, y_j > 0$ for all $i$ and $j$ in any co-occurrence data, a reasonable constraint on modeling independent co-occurrences requires positivity across all joint frequencies: $\min_{ij}\{\hat{X}_{ij}\}>0$. This is evident in marginal-cross-products, for which $\min_{ij}\{\hat{X}_{ij}\} = 1$ due to hapax-legomena ubiquity.

\subsection{Harmonic Statistical Structure}
To avoid the use of \textit{any} data while representing 
a target task's vocabulary, $\mathcal{W}$,
a model of what pre-training \textit{learns} 
is needed---here, 
a distributional model of co-occurrence. 
For documents, marginal distributions of co-occurrences
(unigram distributions) 
can generally be observed to exhibit 
harmonic structure, i.e.,
can generally be modeled via Zipf's law~\cite{zipf1935a,zipf1949a}: $\hat{x}_i = 
N/r_i$. Without loss of generality, the $r_i$, or, \textit{ranks}, intuitively indicate the number of \textit{other} words which occur at least as often. In this presentation, we likewise scale by $N =|\mathcal{W}|$ to ensure the vocabulary's smallest unigram `frequency' is $1$. This should raise a question of alignment---how 
to index the target vocabulary's harmonic structure---
which we resolve by counting and ranking
the target task's training tokens. 
Necessarily, this makes our representation
reliant on \textit{some} empirical information, 
namely an ordering of the target task's 
training data by its vocabulary's ranks ($r_i$).

Now, assuming harmonic unigram frequencies for our model
implies the rows of $\hat{X}_{i,j}$ should marginalize according to $\hat{x}$. 
To model co-occurrences,
we self-sample from $\hat{x}$  
for $2m\hat{x}_i$ other words
to model the sliding window of $\pm m$ words
around each token of the modeled document. 
Since co-occurrences also exhibit hapax legomena,
we set $\min_{ij}\{\hat{X}_{ij}\} = 1$, 
which forces a closed form:
\begin{equation}
    \hat{X}_{ij} = \frac{2mN}{r_ir_jH_N},
    \label{eq:Xhat}
\end{equation}
where $H_N$ is the $N^{\rm th}$ harmonic number.

\subsection{Eigen-Decomposing Distributional Noise}
While there are many matrix factorization methods that could be applied, 
a straightforward approach applies the eigen-decomposition of $\hat{X}$. 
As it turns out, the symmetry of $\hat{X}$ (and any empirical co-occurrence matrix) ensures the existence of a diagonal matrix, $\Lambda$, of unique eigenvalues and an invertible eigen-space matrix, $Q$, that moreover is orthogonal, i.e., with $Q^{-1} = Q^T$. 
This leads to an eigen-decomposition of the form: $\hat{X} = Q\Lambda Q^{T}$. 
This means that the columns of $Q$ are unit vectors---just like a one hot encoding/standard basis set. 
Like with other matrix factorizations, 
a dimensionality reduction to $d < N$ dimensions and approximation of $\hat{X}$ can be derived by the removal of the smallest $N - d$ eigenvalues, $\Lambda_d$.
We retain half of the approximating structure and call it \textit{EigenNoise}.

\begin{table*}[t!]
    \centering
    \hfill
    \begin{tabular}{|c|rr|rr|}
        \hline
        \multicolumn{5}{|c|}{\textbf{Gigaword}} \\
        \hline
        $m$ & \multicolumn{2}{|c|}{\textbf{PoS}} & \multicolumn{2}{|c|}{\textbf{NER}}  \\ \hline
        $0$ & \textbf{88.1} $\pm$ 0.0 & \textbf{88.3} $\pm$ 0.1 & \textbf{92.4} $\pm$ 0.0 & \textbf{92.2} $\pm$ 0.1 \\
        $2$ & \textbf{89.1} $\pm$ 0.0 & \textbf{91.5} $\pm$ 0.0 & \textbf{95.7} $\pm$ 0.1 & \textbf{95.8} $\pm$ 0.1 \\
        $5$ & \textbf{87.2} $\pm$ 0.4 & \textbf{91.1} $\pm$ 0.1 & \textbf{95.4} $\pm$ 0.1 & \textbf{95.6} $\pm$ 0.1 \\
        $10$ & \textbf{85.0} $\pm$ 0.2 & \textbf{90.5} $\pm$ 0.1 & \textbf{94.9} $\pm$ 0.2 & \textbf{95.3} $\pm$ 0.1 \\
        \hline\hline
        \multicolumn{5}{|c|}{\textbf{EigenNoise}} \\ 
        \hline
        $m$ & \multicolumn{2}{|c|}{\textbf{PoS}} & \multicolumn{2}{|c|}{\textbf{NER}}  \\ \hline
        $0$ & 74.2 $\pm$ 0.0 & 86.5 $\pm$ 1.3 & 83.8 $\pm$ 1.2 & 90.3 $\pm$ 0.1 \\
        $2$ & 64.3 $\pm$ 10.4 & 89.5 $\pm$ 0.4 & 87.3 $\pm$ 0.1 & 93.5 $\pm$ 0.1 \\
        $5$ & 71.2 $\pm$ 0.1 & 89.6 $\pm$ 0.2  & 86.9 $\pm$ 0.1 & 93.9 $\pm$ 0.1 \\
        $10$ & 69.0 $\pm$ 0.1 & 89.6 $\pm$ 0.3 & 86.6 $\pm$ 0.1 & 93.7 $\pm$ 0.4 \\
        \hline\hline
        \multicolumn{5}{|c|}{\textbf{Random}} \\ 
        \hline
        $m$ & \multicolumn{2}{|c|}{\textbf{PoS}} & \multicolumn{2}{|c|}{\textbf{NER}}  \\ \hline
        $0$ & 77.1 $\pm$ 0.7 & 81.2 $\pm$ 1.3 & 85.2 $\pm$ 2.8 & 86.8 $\pm$ 1.9 \\
        $2$ & 69.8 $\pm$ 3.1 & 76.7 $\pm$ 1.1 & 84.8 $\pm$ 0.6 & 90.2 $\pm$ 1.4 \\
        $5$ & 63.1 $\pm$ 2.3 & 84.7 $\pm$ 1.6 & 83.2 $\pm$ 1.0 & 91.4 $\pm$ 0.2 \\
        $10$ & 60.0 $\pm$ 0.4 & 85.6 $\pm$ 0.6 & 83.9 $\pm$ 0.2 & 91.6 $\pm$ 0.3 \\
        \hline
    \end{tabular}\hfill
    \begin{tabular}{|c|rr|}
        \hline
        \textbf{Task} & \multicolumn{2}{|c|}{\textbf{Gigaword}} \\
        \hline
        \textbf{I} & \textbf{60.7} $\pm$ 0.6 & \textbf{61.5} $\pm$ 0.8 \\
        \textbf{H} & \textbf{51.3} $\pm$ 0.2 & 51.2 $\pm$ 1.2 \\
        \textbf{O} & \textbf{76.7} $\pm$ 0.6 & \textbf{80.2} $\pm$ 0.7 \\
        \textbf{E} & \textbf{61.2} $\pm$ 0.4 & 66.9 $\pm$ 0.8 \\
        \textbf{S} & 65.7 $\pm$ 5.5 & 64.5 $\pm$ 6.3 \\
        \hline
        \hline
        \textbf{Task} & \multicolumn{2}{|c|}{\textbf{EigenNoise}} \\
        \hline
        \textbf{I} & 51.8 $\pm$ 2.2 & 58.4 $\pm$ 1.5 \\
        \textbf{H} & 47.5 $\pm$ 0.2 & \textbf{52.5} $\pm$ 2.5 \\
        \textbf{O} & 72.9 $\pm$ 0.0 & 76.4 $\pm$ 1.9 \\
        \textbf{E} & 39.7 $\pm$ 0.8 & \textbf{67.6} $\pm$ 1.0 \\
        \textbf{S} & \textbf{66.8} $\pm$ 5.7 & 64.4 $\pm$ 4.0 \\
        \hline
        \hline
        \textbf{Task} & \multicolumn{2}{|c|}{\textbf{Random}} \\
        \hline
        \textbf{I} & 49.1 $\pm$ 2.1 & 52.6 $\pm$ 2.2 \\
        \textbf{H} & 51.2 $\pm$ 1.1 & 50.5 $\pm$ 0.3 \\
        \textbf{O} & 72.9 $\pm$ 0.0 & 72.3 $\pm$ 0.2 \\
        \textbf{E} & 39.5 $\pm$ 0.1 & 41.3 $\pm$ 0.7 \\
        \textbf{S} & 65.7 $\pm$ 4.1 & \textbf{65.6} $\pm$ 4.0 \\
        \hline
    \end{tabular}
    \hfill
    \caption{
        (Left) 
        % Codelength performance 
        Test set accuracy
        on CoNLL2003 tasks.
        Accuracy is averaged across random seeds $\pm$ the standard deviation, with left and right accuracy for frozen and un-frozen embeddings respectively. 
        $m$ indicates the window size.
        (Right) Test set accuracy on TweetEval tasks. Accuracy is averaged across random seeds $\pm$ the standard deviation, with left and right accuracy for frozen and un-frozen embeddings respectively.
    }
    \label{tab:all}
\end{table*}

\section{Experimentation}

To evaluate the performance
of our proposed initialization scheme,
we compare our model 
against a randomly initialized 
(parameters simply drawn from a standard normal distribution)
baseline
as well as empirical GloVe word vectors
trained on the Gigaword corpus \cite{pennington2014glove}.
We evaluate performance on tasks
selected from
two downstream benchmarks:
CoNLL2003 \cite{tjong2003introduction} and TweetEval \cite{barbieri2020tweeteval}.
From CoNLL-2003, 
we consider
Parts-of-Speech (POS) tagging 
and Named Entity Recognition (NER) 
as small-scale, token-based classification tasks
to quantify a baseline ability to represent 
these linguistic constructs in a representation space.
TweetEval is a sequence classification benchmark 
designed to test a model's ability to represent and classify tweets~\cite{barbieri2020tweeteval}.%\footnote{\url{https://github.com/cardiffnlp/tweeteval}}
% something none of these representations
% are specifically trained to achieve. 
We select 5 of the 7 sub-tasks to
explore regularity %of representation
in social labels:
irony (\textbf{I}), % \cite{van2018semeval},
hate speech (\textbf{H}), % \cite{basile2019semeval},
offensive language (\textbf{O}), % \cite{zampieri2019semeval},
emotion (\textbf{E}), % \cite{mohammad2018semeval},
and stance (\textbf{S}). % \cite{mohammad2016semeval}.
% TweetEval is freely available,
% and the Twitter data is anonymized to remove usernames.
% Given the nature of these tasks,
% some of their content is offensive.

\section{Results \& Discussion}

\subsection{CoNLL}

% \textbf{CoNLL:}
Table \ref{tab:all} (Left) presents the results of probing
on CoNLL-2003.
% Results are averaged across random seeds,
% with the results of various window sizes $w$ and 
% the effects of freezing and un-freezing the embedding layer 
% of the model shown.
% From this Table, several broad trends emerge.
Consistently, backpropagating through representations
reduces the codelength.
This isn't surprising;  %as, by far,
the embedding layer %of the probe 
contains the most parameters. 
However, what is surprising is that % the fact that 
EigenNoise starts at high codelengths
(indicating poor regularity with respect to the labels),
but, when allowed to update,
is able to approach the codelengths of empirical GloVe.
This suggests that, while EigenNoise
isn't quite ideal immediately, % out of eigendecomposition,
if allowed to adapt to the task at hand,
it can do so with relatively little data.
When factoring in the naivety of EigenNoise,
the fact that it can approach the empirical GloVe model 
that has a far larger vocabulary (400K words versus 20K ranks)
and is trained on infinitely more data,
these result are more compelling. Other interesting observations include that the theory-based vectors
do worse compared to the standard normal random vectors
when both are held static. 
However, when both are allowed to update their representations,
the random vectors barely reduce the codelength
whereas the theory-based vectors more than halve theirs.
At the very least, this indicates
the theory-based rank vectors are an interesting weight initialization point.

\subsection{TweetEval} 

% \textbf{TweetEval.} 

Table \ref{tab:all} (Right) displays the results
of probing on TweetEval.
Here, we observe that the random vectors 
are clearly the worst overall,
but that all these representations perform similarly for these types of tasks,
with the empirical GloVe model performing best out-of-the-box.
This seems fairly reasonable, given the way each model was constructed. One may also observe that the empirical and theory vectors
result in similar codelengths for the hate speech detection 
and offensive language identification tasks
when the theory vectors are allowed to update. 
This seems to indicate that the theory-based vectors 
do not contain the regular signal needed to detect these social phenomena initially but that the empirical GloVe vectors do. 
Seemingly, empirically-based GloVe vectors contain a higher degree 
of information about hate speech and offensive language out-of-the-box 
when compared to an EigenNoise set that's free from such biases,
yet the latter can adapt through tuning.

\section{Conclusion \& Future Work}

In this work,
we introduce an incredibly naive 
initialization scheme for independent word vectors such as GloVe and word2vec.
We provide preliminary experimentation
that demonstrates
the efficacy of such a scheme
in a low-compute setting
through an information-theoretic approach 
with MDL probing.
We believe that these preliminary results
are interesting 
and beg further investigation,
especially as an initialization scheme
for independent word vectors 
even if they are to be empirically tuned.

\bibliography{anthology,custom}
\bibliographystyle{acl_natbib}

\appendix

\begin{table*}[t!]
    \centering
    \hfill
    \begin{tabular}{|c|rr|rr|}
        \hline
        \multicolumn{5}{|c|}{\textbf{GigaWord}} \\
        \hline
        $m$ & \multicolumn{2}{|c|}{\textbf{PoS}} & \multicolumn{2}{|c|}{\textbf{NER}}  \\ \hline
        $0$ & \textbf{85.5} $\pm$ 9.5 & \textbf{85.4} $\pm$ 0.3 & \textbf{47.2} $\pm$ 0.1 & \textbf{47.9} $\pm$ 0.4 \\
        $2$ & \textbf{99.2} $\pm$ 0.4 & \textbf{79.7} $\pm$ 0.5 & \textbf{32.7} $\pm$ 0.4 & \textbf{29.4} $\pm$ 0.1 \\
        $5$ & \textbf{121.4} $\pm$ 0.7 & \textbf{91.6} $\pm$ 0.4 & \textbf{38.5} $\pm$ 0.3 & \textbf{33.5} $\pm$ 0.6 \\
        $10$ & \textbf{142.4} $\pm$ 1.2 & 104.4 $\pm$ 0.3 & \textbf{44.8} $\pm$ 0.6 & \textbf{38.6} $\pm$ 0.3 \\
        \hline\hline
        \multicolumn{5}{|c|}{\textbf{EigenNoise}} \\ 
        \hline
        $m$ & \multicolumn{2}{|c|}{\textbf{PoS}} & \multicolumn{2}{|c|}{\textbf{NER}}  \\ \hline
        $0$ & 221.9 $\pm$ 4.5 & 110.8 $\pm$ 0.1 & 121.7 $\pm$ 1.5 & 66.4 $\pm$ 0.2 \\
        $2$ & 205.1 $\pm$ 1.7 & 90.9 $\pm$ 0.3 & 89.7 $\pm$ 0.6 & 40.5 $\pm$ 1.0 \\
        $5$ & 218.6 $\pm$ 0.5 & 92.4 $\pm$ 0.2  & 91.3 $\pm$ 0.2 & 41.8 $\pm$ 0.6 \\
        $10$ & 239.0 $\pm$ 0.6 & \textbf{96.9} $\pm$ 0.5 & 97.1 $\pm$ 0.2 & 44.7 $\pm$ 1.1 \\
        \hline\hline
        \multicolumn{5}{|c|}{\textbf{Random}} \\ 
        \hline
        $m$ & \multicolumn{2}{|c|}{\textbf{PoS}} & \multicolumn{2}{|c|}{\textbf{NER}}  \\ \hline
        $0$ & 157.8 $\pm$ 3.8 & 137.7 $\pm$ 2.9 & 95.1 $\pm$ 1.8 & 83.7 $\pm$ 0.7 \\
        $2$ & 197.7 $\pm$ 12.4 & 129.3 $\pm$ 1.4 & 103.5 $\pm$ 3.3 & 62.4 $\pm$ 1.0 \\
        $5$ & 252.5 $\pm$ 4.8 & 138.2 $\pm$ 1.8 & 116.7 $\pm$ 2.2 & 65.4 $\pm$ 2.1 \\
        $10$ & 281.9 $\pm$ 10.9 & 147.1 $\pm$ 1.7 & 125.3 $\pm$ 1.8 & 69.7 $\pm$ 1.6 \\
        \hline
    \end{tabular}\hfill
    \begin{tabular}{|c|rr|}
        \hline
        \textbf{Task} & \multicolumn{2}{|c|}{\textbf{Gigaword}} \\
        \hline
        \textbf{I} & \textbf{1.9} $\pm$ 0.0 & \textbf{1.9} $\pm$ 0.0 \\
        \textbf{H} & \textbf{5.3} $\pm$ 0.1 & \textbf{5.0} $\pm$ 0.1 \\
        \textbf{O} & \textbf{6.7} $\pm$ 0.0 & \textbf{6.5} $\pm$ 0.1 \\
        \textbf{E} & \textbf{3.3} $\pm$ 0.0 & \textbf{3.1} $\pm$ 0.0 \\
        \textbf{S} & \textbf{0.5} $\pm$ 0.1 & \textbf{0.5} $\pm$ 0.1 \\
        \hline
        \hline
        \textbf{Task} & \multicolumn{2}{|c|}{\textbf{EigenNoise}} \\
        \hline
        \textbf{I} & \textbf{1.9} $\pm$ 0.0 & \textbf{1.9} $\pm$ 0.0 \\
        \textbf{H} & 5.9 $\pm$ 0.0 & \textbf{5.0} $\pm$ 0.0 \\
        \textbf{O} & 7.5 $\pm$ 0.0 & 6.8 $\pm$ 0.1 \\
        \textbf{E} & 4.1 $\pm$ 0.0 & 3.5 $\pm$ 0.0 \\
        \textbf{S} & \textbf{0.5} $\pm$ 0.1 & \textbf{0.5} $\pm$ 0.1 \\
        \hline
        \hline
        \textbf{Task} & \multicolumn{2}{|c|}{\textbf{Random}} \\
        \hline
        \textbf{I} & 2.0 $\pm$ 0.0 & 1.9 $\pm$ 0.0 \\
        \textbf{H} & 6.0 $\pm$ 0.0 & 5.7 $\pm$ 0.1 \\
        \textbf{O} & 7.5 $\pm$ 0.0 & 7.5 $\pm$ 0.0 \\
        \textbf{E} & 4.1 $\pm$ 0.0 & 4.1 $\pm$ 0.0 \\
        \textbf{S} & \textbf{0.5} $\pm$ 0.1 & \textbf{0.5} $\pm$ 0.1 \\
        \hline
    \end{tabular}
    \hfill
    \caption{
        (Left) Codelength performance on CoNLL2003 tasks, measured in kilobytes.
        Codelengths are averaged across random seeds $\pm$ the standard deviation, with left and right codelengths for frozen and un-frozen embeddings respectively. 
        $m$ indicates the window size.
        (Right) Codelength performance on TweetEval tasks. Codelengths are measured in kilobits. Codelengths are averaged across random seeds $\pm$ the standard deviation, with left and right codelengths for frozen and un-frozen embeddings respectively.
    }
    \label{tab:all-mdl}
\end{table*}

\section{Experimental Details for Probing}
The probes used in this work are simple multi-layer perceptrons
with a single hidden layer, hidden dimension of 512, and no dropout, defined as: $\hat{y}_i \sim \text{softmax}(W_2 \text{ReLU}(W_1 h_i))$.
% Specifically, we use a hidden dimension of 512 and no dropout. 
For sequence classification tasks, the entire sequence is embedded and then averaged.
For token classification tasks, a window size $m \in \{0, 2, 5, 10\}$ is selected and the $2w+1$ token window is embedded and flattened.
Each experiment is repeated 3 times on random seeds $\in \{0, 1234, 322111\}$ with data block splits chosen to align with previous work: 
0.1, 0.2, 0.4, 0.8, 1.6, 3.2, 6.25, 12.5, 25, 50, 100 \% of the data.

\subsection{Representations} 
We compare three representations: 
GloVe trained on the GigaWords corpus \cite{pennington2014glove},
our EigenNoise model,
and a baseline where parameters are sampled from a standard normal distribution.
% each embedding parameter is drawn from a standard normal distribution. 
EigenNoise uses a vocab size of $N=20,000$,
just large enough to fit the training vocab of each data set
to demonstrate the efficacy of this approach in a ``low-compute'' setting.
% and exposed to infinitely less data than the empirical GloVe models.
For all representations, a dimensionality of 50 is used
and both freezing and un-freezing the embedding layer is explored.
% Backpropagating through the embedding layer 
% produces an online codelength that quantifies 
% how easy it is to adapt the initial embedding space 
% to encode the desired properties.

\subsection{Optimization} 
All probes are trained
with the Adam optimizer \cite{kingma2015adam} 
with an initial learning rate 0.001. 
Adhering to previous works \cite{hewittliang2019designing,voita2020information}, 
we anneal the learning rate by a factor of 0.5 once the epoch
does not lead to a new minimum loss on the development set; training stops after 4 such epochs.

\subsection{Hardware} 
Experiments were completed using a single NVIDIA Titan V 12GB on our internal cluster.
The combination of representations, heterogeneous dataset,
and early stopping criteria result in variable length runs,
however, the longest single probe run took no more than 2 hours to complete.

\section{Dimensionality Reduction}
To precisely compute the eigen-decomposition dimensionality reduction, define $I_d\in\mathbb{R}^{N\times d}$ to be the first $d$ columns of the $N$-dimensional identity matrix ($I$) and let $\Lambda_d\in\mathbb{R}^{d\times N}$ denote the first $d$ rows of the diagonal eigenvalue matrix. The $\hat{X}$-reconstruction equation is then:
\begin{equation}
\hat{X}\approx QI_d (Q\Lambda_d)^T = U_dV_d^T    
\label{eq:EigenNoise}
\end{equation}
where $U_d = QI_d$ and $V_d = Q\Lambda_d$ are needed to retain the effect of  zeroing out $\Lambda$'s $N-d$ smallest diagonal elements. This reduces the $Q$-variation into two low-dimensional ($d$) representations that approximately reconstruct $\hat{X}$. 
For our purposes, we retain $U_d$
and refer to the solution as \textit{EigenNoise}.

We note that varying choices could be made to handle $\Lambda$---it could be multiplied without loss of generality into the $U$-side, instead of the $V$-side. But perhaps more interestingly, $\Lambda$'s values could be rooted---perhaps over $\mathbb{C}$---for a symmetric set, i.e., with $U_d = V_d$ and $U_d, V_d\in\mathbb{C}^{N \times d}$. We speculate that informative variation over $\mathbb{C}$ may exist, but leave the exploration of this to future work.

\section{Information-Theoretic Evaluation}
Here, we adopt an alternate, 
information-theoretic probing methodology for evaluation
% that is designed to incorporate not only 
that combines the measure of ease of mapping 
from representation to label space
as well as the complexity of the model needed to do so.
% a representations ability to be mapped from representation space to labels space
% but also incorporates the complexity of the model needed to do so.
This method, called Minimum Description Length (MDL) \cite{voita2020information}, 
% is an information-theoretic approach to probing a representation that 
is concisely described as measuring the regularity of a representation 
with respect to a set of labels. 
Specifically, we adopt the online codelength metric (measured in kilobits), 
where a smaller codelength is indicative of a more regular representation.
We adopt this metric as it is 
% As demonstrated in the original work on MDL \cite{voita2020information},
% this metric is
more informative than accuracy 
and is more stable with respect to random initializations and hyperparameter selection.

\subsection{MDL Probing}

As discussed in the related works,
comparing the performance of pre-trained representations
can be more subtle than simply training a classifier (i.e., a \emph{probe}) 
and comparing the attained performance,
sometimes giving un-intuitive results 
such as random baselines performing comparably well to pre-trained ones. 
To combat this issue, 
we adopt the information-theoretic approach 
of Minimum Description Length (MDL) probing
\cite{voita2020information},
which serves as a measure 
of the regularity of a representation
with respect to a label set.
This allows us to quantify
how much difficulty a classifier has
in achieving a particular level of performance.

In MDL probing, let 
$$\mathcal{D} = \{(x_1, y_1), (x_2, y_2), ..., (x_n, y_n)\}$$
be a dataset 
where $x_{1:n}=(x_1, x_2, ..., x_n)$ are representations from a model 
and $y_{1:n}=(y_1, y_2, ..., y_n)$ are the labels of a desired property.
Instead of measuring how well a probe can perform this mapping,
MDL tasks a probe with learning to efficiently transmit the data using the representation.
Using the online codelength metric, assume that two agents (Alice and Bob)
agree upon a form of a model $p_{\theta}(y|x)$ with learnable weights $\theta$, 
a random weight initialization scheme,
and an optimization procedure.

Break points $1 = t_0 < t_1 < ... < t_S = n$ are selected
to form data blocks to be transmitted. Alice begins by transmitting $y_{1:t_1}$ using a uniform code, 
from which both Alice and Bob train a model $p_{\theta_1}(y|x)$
using the first data block $\{(x_i, y_i)\}_{i=1}^{t_1}$.
Alice uses that model to transmit the next data block $y_{t_1 + 1:t_2}$,
which is used to train a better model $p_{\theta_2}(y|x)$ to transmit the next block.
This continues until all data has been transmitted,
resulting in an online codelength computed via 
 $L^{\text{online}}(y_{1:n}|x_{1:n}) = t_1\log_2{K} - \sum_{i=1}^{S-1}\log_2{p_{\theta_i}(y_{t_{i} + 1:t_{i+1}}|x_{t_{i} + 1:t_{i+1}})}$.
As in \cite{voita2020information},
probes that learn mappings via fewer 
data points will have shorter codelengths.

% \section{Example Appendix}
% \label{sec:appendix}

% This is an appendix.

\end{document}